\documentclass[]{spie}  %>>> use for US letter paper
\usepackage{amsmath,amsfonts,amssymb}
\usepackage{graphicx}
\usepackage[colorlinks=true, allcolors=blue]{hyperref}

\title{PROB-SLAM: Real-time Visual SLAM Based on Probabilistic Graph Optimization}
\author{Xianwei Meng , Bonian Li}
\authorinfo{Further author information: (Send correspondence to Xianwei Meng)\\Xianwei Meng: E-mail: 2606910919@qq.com\\Bonian Li: E-mail: 584119429@qq.com, Telephone: +86 139 1004 4790}
\begin{document} 
\maketitle

\begin{abstract}
Traditional SLAM algorithms are typically based on artificial features, which lack high-level information. By introducing semantic information, SLAM can own higher stability and robustness rather than purely hand-crafted features. However, the high uncertainty of semantic detection networks prohibits the practical functionality of high-level information. To solve the uncertainty property introduced by semantics, this paper proposed a novel probability map based on the Gaussian distribution assumption. This map transforms the semantic binary object detection into probability results, which help establish a probabilistic data association between artificial features and semantic info. Through our algorithm, the higher confidence will be given higher weights in each update step while the edge of the detection area will be endowed with lower confidence. Then the uncertainty is undermined and has less effect on nonlinear optimization. The experiments are carried out in the TUM RGBD dataset, results show that our system improves ORB-SLAM2 by about 15\% in indoor environments’ errors. We have demonstrated that the method can be successfully applied to environments containing dynamic objects.
\end{abstract}
% Include a list of keywords after the abstract 

\section{INTRODUCTION}
\label{sec:intro}  % \label{} allows reference to this section

Over the past few decades, with the development of science and technology, the use of computers for simultaneous localization and mapping (SLAM) has become an indispensable part of many automation fields. It enables the robot to have the ability to locate and obtain the location information of the environment in a new environment, so it is a crucial technology for autonomous navigation of mobile robots\cite{1} and is widely used in autonomous driving\cite{2}, drones and augmented reality (AR) \cite{3} and other fields.

In the past few decades, the SLAM problem has been a widely used research direction, so many excellent visual SLAM systems have emerged, such as ORBSLAM\cite{4}, ORB-SLAM2\cite{5}, LSD-SLAM\cite{6}, and SVO\cite{7}. When the environment is static, or there are few dynamic elements, these excellent systems can achieve satisfactory performance. However, unstable environmental factors are ubiquitous, and sometimes the movement of people in indoor positioning will affect the positioning results. Huge destruction. Due to static world assumptions and lighting or other influences, the above standard SLAM systems are significantly less accurate and less robust in such highly dynamic environments.

An important branch of SALM, dynamic SLAM algorithms, can be a hot research area today. This technology can already be used in many real-world applications, such as mobile robotics. For traditional dynamic SLAM algorithms, Cadena et al. provide a more comprehensive overview of the past, present, and future of SLAM\cite{8}. However, this type of dynamic SLAM has no prior semantic information. In the absence of semantic information as a priori, using reliable constraints to find the right feature matching relationship is the primary method to deal with dynamic SLAM problems. Li et al. proposed a static weighting method for keyframe edge points and integrated it into the IAICP method to reduce tracking error\cite{9} Jin et al. proposed a dense visual odometry calculation method based on a background model, estimated from the depth scene Nonparametric Background Model\cite{10}. Dai et al. distinguish dynamic and static map points based on feature correlation\cite{11}. Flow fusion uses optical flow residuals to highlight dynamic regions in RGB-D point clouds\cite{12}. Since deep learning networks do not need to provide semantic priors, the above methods are generally fast in handling dynamic factors but lack accuracy. In response to this problem, VSO\cite{13} proposed a novel method. If the probability of each object in the picture is obtained based on the semantic segmentation of a particular frame, that is, the probability of using the object is inversely proportional to the distance from the center of the object, The concept of distance can be used to create the concept of normalization, which reduces the background part that is far away in semantic segmentation, thereby improving the accuracy of pose estimation, but because the error of semantic segmentation itself is more prominent, and semantic segmentation often There are gaps between different instances, resulting in lower overall accuracy, so this method is not very helpful for practical results.

Specifically, in deep learning methods, the most common techniques in the deep learning section\cite{14} are usually adopted to generate the necessary semantic information, such as object detection and semantic segmentation techniques. Li et al. used the well-known semantic segmentation network SegNet to segment images, which were further processed to distinguish dynamic objects\cite{15}. Zhang et al. adopted YOLO\cite{16} to identify objects in the environment and build a semantic map to filter dynamic feature points to improve the system's accuracy. Liu et al. adopted different semantic segmentation methods to detect dynamic objects and remove outliers\cite{17}. Deep learning methods are helpful when dynamic objects in the environment are known in advance. However, deep learning methods rely heavily on the quality of the network\cite{18}. A network with a simple architecture may be unable to effectively identify dynamic objects in some cases, while a complex network architecture may slow down the system.

A subset of SLAM also applies CNNs to segment previously dynamic objects in images and re-examine them using multi-view geometric models. Zhao et al. \cite{19} adopted a Mask-RCNN network to discover the contours of latent dynamic objects. They then checked the consistency of optical flow to detect the actual state of dynamic regions. Yang et al. \cite{20} first used an object detection model to detect predefined dynamic objects and then adopted multi-view constraints to confirm dynamic pixel depth images. However, semantic segmentation, such as SegNet and Mask-RCNN, is time-consuming, and multi-view methods require multiple frames of images. This makes the SLAM system face practical and stability challenges in practical applications.

In detail, SLAM generally has two basic requirements: robustness of tracking and real-time performance. Therefore, detecting dynamic objects that populate the scene and preventing tracking algorithms from using data associations related to these dynamic objects in real-time is a challenge that enables SLAM to be deployed in the real world. With the development of deep learning technology, many studies have introduced semantic information as a constraint to solve dynamic problems in SLAM. By means of object recognition and semantic segmentation, constraints are added to the feature point information distributed on objects. SLAM performance tends to be significantly improved. First, Ryan R et al. proposed the concept of probabilistic semantics. They added geometry, semantics, and sensors into the same framework for optimization and solved it with the EM algorithm, realizing a SLAM system with higher positioning accuracy. The sensor and the lens are used to observe the correspondence between the road signs and the road signs, and the weight factor w represents it. The probability model calculates the object's center, and then the second image projection is performed to approach the center of the detection frame. Konstantinos-Nektarios Lianos et al. \cite{13} proposed a VSO algorithm. They proposed to use semantic segmentation to obtain the probability of each object in the picture. However, the boundary of semantic segmentation often lacked accuracy, and the real-time performance was also poor. Poor and these defects are more significant when there are dynamic objects in the scene, so the practice of introducing semantic information into SLAM cannot be used in most scenes with dynamic objects.

In response to the above problems, we tightly organize deep learning and static SLAM. At the same time, the two-dimensional Gaussian distribution is used to process the probability distribution map, which effectively reduces the negative impact of inaccurate semantic segmentation borders, realizes the filtering of feature points distributed on dynamic objects, and retains sufficient information on static objects to the greatest extent. Feature points, to adjust the weight of feature points distributed on dynamic objects in optimization so that the dynamic SLAM system is more stable and less affected by moving objects - we propose PROB-SLAM, which is based on a High-performance, visual SLAM system for object detection and probability distribution maps in dynamic indoor environments. Based on ORB-SLAM2\cite{5}, this Prob-SLAM system uses YOLO\cite{16} to identify dynamic objects based on predefined knowledge, obtains semantic information through YOLOv5, and distinguishes high-confidence regions from the boundary range obtained by semantic segmentation. In the low-confidence region, Gaussian distribution is introduced into the high-confidence region, a probability distribution map is finally constructed, SLAM is optimized from a new dimension, and the attribute of confidence of critical points is given according to the probability distribution map under various constraints. Next, calculate the feature point information required for pose estimation. The confidence of the critical points is used as a weight to participate in the camera pose optimization. In this way, the goal of reducing the importance of dynamic objects in SLAM and retaining the information of static objects to the greatest extent is achieved, which makes good use of semantic information and reduces the negative impact of inaccurate semantic segmentation. To verify the effectiveness of PROB-SLAM, we conduct dynamic monocular SLAM experiments on the TUM dataset. Our method achieves higher-than-original localization accuracy in both low-dynamic and high-dynamic scenes on the same dataset compared with the original ORB-SLAM. The main contributions of this paper are as follows:

\noindent\textbf{1)} A probabilistic algorithm based on the foreground (within the bounding box) and background (outside the bounding box) information is attached to the key points. Aiming at the error problem caused by dynamic objects, a PROB algorithm based on semantic segmentation is proposed to weaken the influence of dynamic objects and preserve the influence of static objects.

\noindent\textbf{2)} Based on the YOLOv5 target detection network and predefined knowledge, distinguish between dynamic objects and static objects, distinguish the picture frame's foreground and background points, and perform a probability algorithm based on two-dimensional Gaussian distribution for the foreground points to minimize the inaccurate boundary of semantic segmentation. With the coming influence, the final construction gets the probability distribution map.

\noindent\textbf{3)} The probability distribution map constructed according to the target motion attribute weakens the optimization weight of the dynamic object, provides it to the subsequent optimization algorithm and processes it in the subsequent newly added PROB thread, and uses the probability distribution map to adjust the covariance matrix of each error term in the nonlinear optimization process, to improve the robustness and adaptability of SLAM systems.

\section{Proposed Method}
This section will detail the overall architecture of PROB-SLAM proposed for dynamic environments. It consists of three threads running in parallel: the object detection thread, the probability distribution map processing thread, and the improvement tracking thread. The object detection thread is responsible for scene semantic segmentation, the probability distribution map processing thread is responsible for the construction of the confidence distribution map, and the improved tracking thread is responsible for the construction of the BA problem algorithm based on the confidence distribution map:

   \begin{figure} [ht]
   \begin{center}
   \begin{tabular}{c} %% tabular useful for creating an array of images 
   \includegraphics[height=5.5cm]{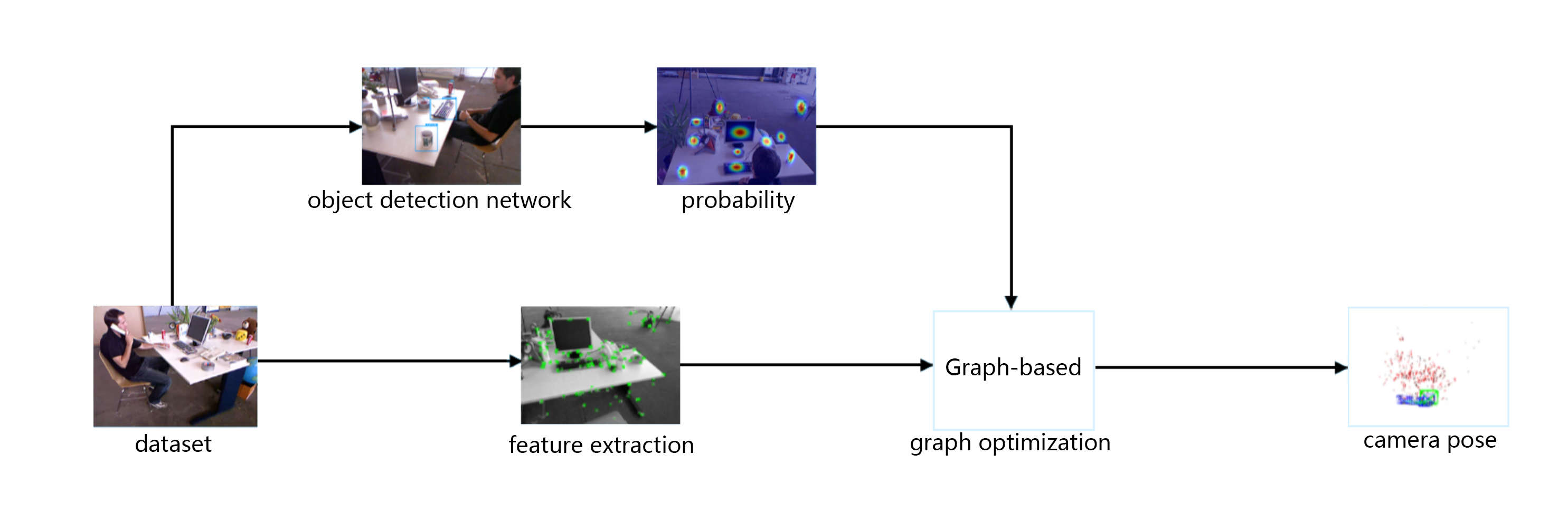}
   \end{tabular}
   \end{center}
   \caption[example] 
   { \label{fig:PM1} 
   RGB images are processed by two independent threads. In the first thread, RGB images are directly subjected to feature extraction, while in the second thread, RGB images are first detected by yolov5, and then a probability distribution map is constructed according to the PROB algorithm. The feature points extracted by one thread and the probability distribution map obtained by the second thread will be transmitted as a topic respectively, and then the pose estimation will be performed in the backend based on those feature points and probability distribution map.}
   \end{figure} 

\subsection{Scene Semantic Segmentation}
In this paper, semantic segmentation is defined as identifying the content and position of each object in the scene. This paper mainly focuses on water cups, display screens, and other objects that remain stationary in space. These objects are stationary, and their shapes contain more features, such as sharper edges or corners. These objects contain more feature points, and because these objects are stationary, the feature points on these objects are more stable than other feature points. We believe that the feature points contained in these stationary objects can maintain better stability after the camera moves. Therefore, it is more reliable for localization and camera pose estimation, and dynamic objects such as people tend to exacerbate the uncertainty brought by semantic segmentation, so we directly regard dynamic objects as part of the background. We assign higher confidence to the feature points on these particular stationary objects to reduce the probability of feature mismatch and the computational burden. We construct a set of static objects that we focus on based on predefined knowledge and use yolov5 according to the set of static objects. Perform scene semantic segmentation in this paper.

\subsection{Construction of confidence distribution map}
We construct a confidence distribution map based on the position of the center point of the static object obtained by scene semantic segmentation. We try two distributions, normal and Gaussian distributions, and compare the robustness and accuracy of the two distributions to our method. Select Gaussian as the distribution. Introducing a Gaussian distribution into the bounding box can effectively alleviate the negative impact of inaccurate semantic segmentation. The two-dimensional Gaussian distribution can minimize the impact of non-target static objects within the bounding box on subsequent pose estimation. Influences. We take the center of the stationary object obtained from the scene semantic segmentation as the center of the two-dimensional Gaussian distribution, assign the center of the two-dimensional Gaussian distribution with a confidence of 0.99, and assign the edge of the two-dimensional Gaussian distribution and the area outside the edge with a confidence of 0.1. The calculation is the same as the size of the scene, and each pixel has a confidence distribution map of its corresponding feature points; that is, a confidence attribute ${prob}_{i,j}$ is assigned to each point $p_{i,j}$

\begin{equation}
\label{eq:fov}
{prob}_{i,j}=\left\{
   \begin{array}{lr}
   \frac{1}{{2\pi}^\frac{d}{2}\varepsilon^\frac{1}{2}}\exp[-\frac{1}{2}\left(X-u)^T\varepsilon^{-1}\left(X-u\right)\right],ifintheboundingbox &  \\
   &  \\
   0.1,ifouttheboundingbox &  
   \end{array}
\right.
\end{equation}

Among them, $\varepsilon$ is the covariance matrix describing the correlation between variables, d is the variable dimension, u is the center coordinate vector of the semantic segmentation box, and $X$=$(i, j)$ is the coordinate vector of the point $p_{i,j}$ ,$ i $and $j$ are the x and y-axis coordinates of the pixel's position in the screen, respectively, so ${prob}_{i,j}$ is calculated according to the pixel's position on the image. Suppose the same pixel appears in multiple In the case of the bounding box, according to our comparative experiments. In that case, we finally select the maximum value of the probability values calculated in each bounding box as the probability value of the point.

\subsection{BA Problem Algorithm Based on Confidence Distribution Graph}
In the SLAM process, Xiang Gao et al.\cite{23} proposed that, it is usually necessary to infer the pose and positioning according to the noise due to various interference and misrecognition noise. According to the confidence distribution map, we assign confidence to each pixel of each frame of the picture. Attribute ${prob}_{i,j}$, when the feature point extraction is completed, and we can optimize the back-end according to the confidence attribute of the feature point. Since SLAM uses G2O optimization to solve the Bundle Adjustment problem, we multiply the probability attribute into the covariance matrix corresponding to each error term of the least squares problem constructed by the Bundle Adjustment problem to introduce the probability attribute to each error term. Weights are adjusted.

\begin{equation}
\label{eq:fov}
T^\ast=\arg{min{\frac{1}{2}}}\sum_{k=1}^{n}{{prob}_k\ast||}U_k-\frac{1}{s_k}KTP_k||_2^2 \, ,
\end{equation}

$T$ is the Lie group of the camera pose $R$, $t$, $u$ is the projection coordinate of a particular spatial point $P_k={X_k, Y_k, Z_k}$,$ U_k={u_k, v_k}^T$, $K$ is the camera internal parameter matrix, in the g2o optimizer In, we control the weight of the edge to achieve the same purpose of controlling the weight of the error term, to realize the prob-based BA problem-solving method.

   \begin{figure} [ht]
   \begin{center}
   \begin{tabular}{c} %% tabular useful for creating an array of images 
   \includegraphics[height=11cm]{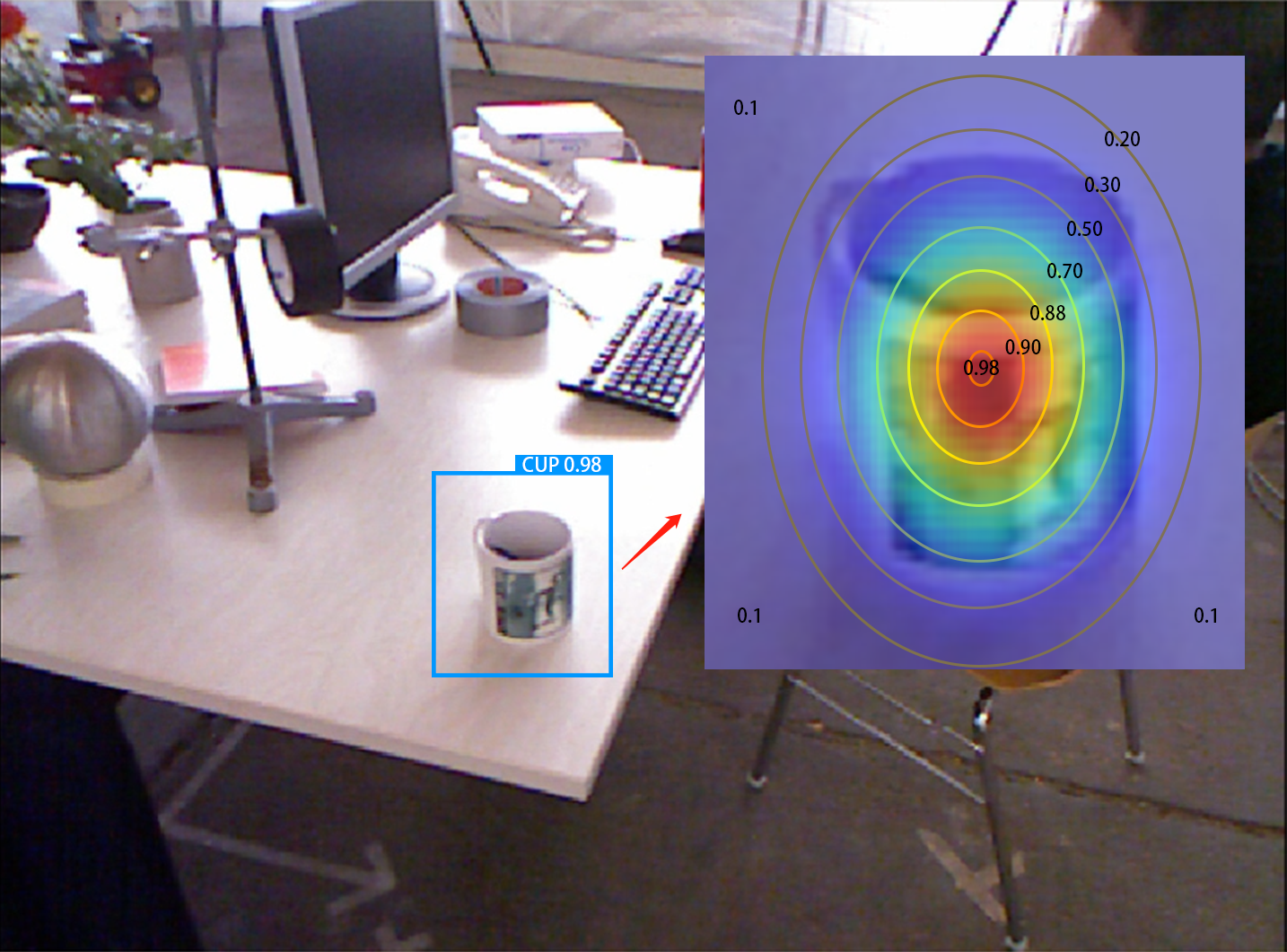}
   \end{tabular}
   \end{center}
   \caption[example] 
   { \label{fig:PM2} 
   This picture shows the local working principle of the probability map. The target detection network captures objects for identification and defines the objects in the form of Gaussian distribution. The center position is defined by the confidence of the target detection object. The probability map gradually reduces the probability as the object's center radiates to the outer circle. The numbers in the picture are the weight of the approximate position of the object.}   
   \end{figure} 

\section{Specific Implementation}

\subsection{Target detection optimizes}
This experiment is based on the ROS system for data transmission. In our scheme, on the one hand, a lightweight network, yolov5, is used for semantic segmentation. On the other hand, the positioning function of the target detection network is used to obtain information related to the target recognition object. For example, the bounding box's center coordinate information and the bounding box's side length information make target detection a crucial auxiliary tool for SLAM positioning. To alleviate the impact of the target detection network on the timeliness of trying SLAM as much as possible, we choose YOLOv5 for target detection. We determine the set of dynamic and static objects that may appear in the scene according to the environment and prerequisites. Based on this target recognition. After obtaining the recognition result, we perform semantic segmentation on the original image and finally obtain the bounding box's center position and side length information.

\subsection{Probability graph thread}
Object detection techniques have great advantages in real-time performance compared with purely semantic segmentation methods but cannot provide accurate object masks. In the indoor dynamic SLAM scene, by setting up a probability distribution map module after object detection, we reduce the influence of these feature points in the subsequent process by assigning low confidence to each feature point distributed in the non-target recognition object. At the same time, according to the two-dimensional Gaussian distribution formula, the confidence of the feature points in each bounding box is defined for the bounding box information obtained in the semantic segmentation, and the attention to the particular static objects existing in the environment is strengthened. The distribution is distributed on these particular static objects. The feature points are more involved in optimization and pose estimation, which enhances positioning accuracy. At the same time, the two-dimensional Gaussian distribution is used because it can reduce the negative impact of inaccurate semantic segmentation as much as possible, thereby improving the effect of camera pose estimation, improving the accuracy of camera pose estimation, and reducing the amount of computation. Finally, we construct a confidence matrix and transmit it as a new topic according to the confidence prob attribute obtained from the two-dimensional Gaussian distribution.

\subsection{Optimization in G2O}
After ORB-SLAM2 normally processes the lens image information from the monocular camera, we combine the probability distribution map with the original static SLAM and multiply the probability attributes of each positioning point into the Bundle Adjustment problem constructed by the Bundle Adjustment problem. The covariance matrix corresponds to each error term of the good positive assist.

\section{Experimental Evaluation}
In this section, we use TUM's RGB-D dataset\cite{21}, which has more dynamic objects, to test the performance optimization of our proposed algorithm for SLAM that introduces semantic segmentation methods. The metrics used to evaluate the accuracy are absolute trajectory error (ATE) and relative pose error (RPE). ATE stands for global consistency of trajectories. RPE includes translation drift and rotation drift. The root means square error (RMSE) and standard deviation (S.D.) of both are used to represent the robustness and stability of the system\cite{22}.

We first show the impact of the PROB algorithm on the positioning points of the original SLAM to detect whether the PROB algorithm will cause too much performance impact on the original ORB-SLAM2\cite{5} and then compare our method with the original ORB-SLAM2 running map trajectory, To determine whether the error has been reduced, to achieve the optimal effect, and then design a series of ablation experiments to test the influence of the PROB module, as well as the natural effect of the probability map concept created by running it alone, and finally conduct real-time analysis and calculation. The size of the impact on time efficiency. All experiments were performed on a computer with an Intel i7 CPU, Nvidia 1070 GPU, and 16GB of memory.

   \begin{figure} [ht]
   \begin{center}
   \begin{tabular}{c} %% tabular useful for creating an array of images 
   \includegraphics[height=8cm]{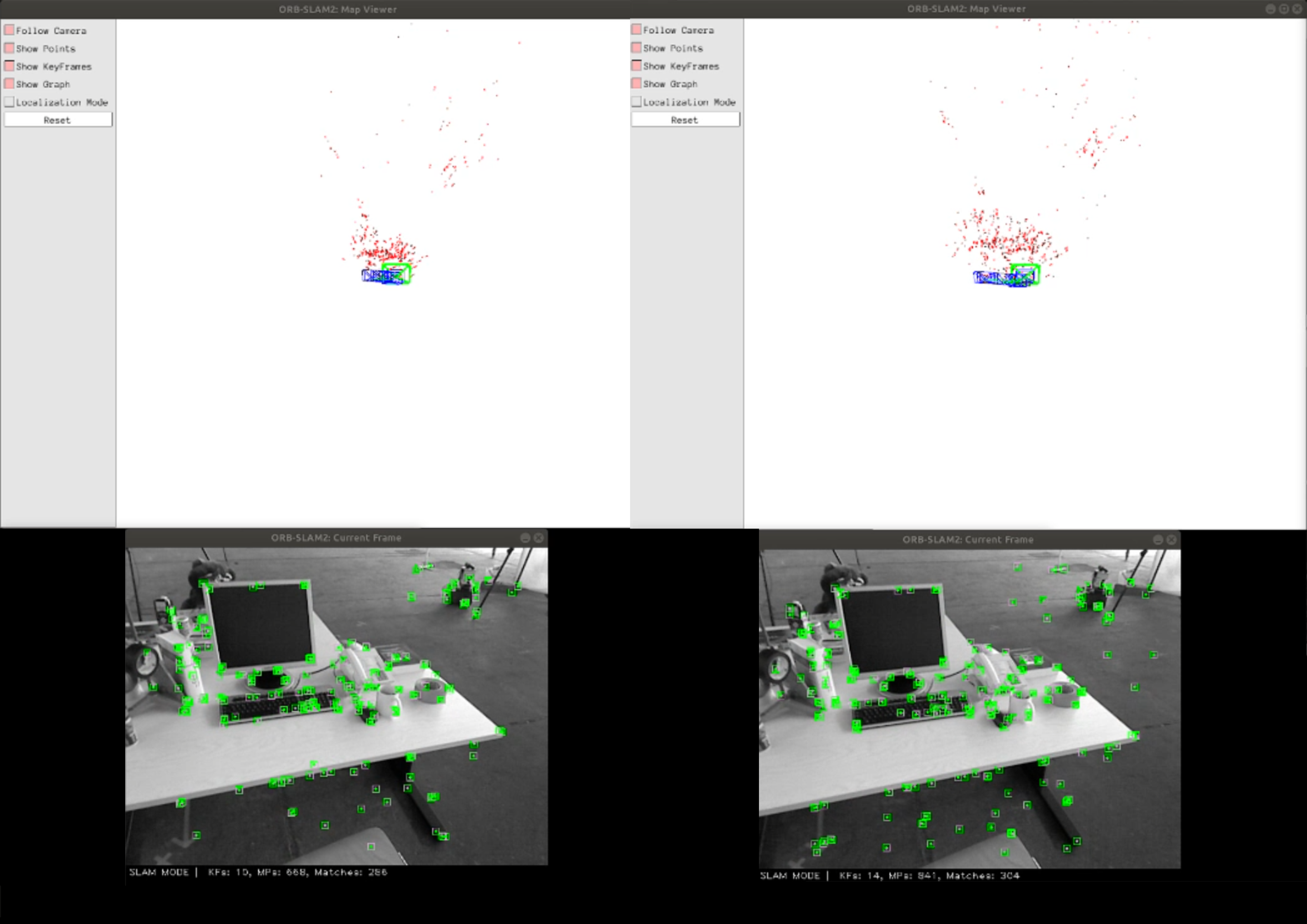}
   \end{tabular}
   \end{center}
   \caption[example] 
%>>>> use \label inside caption to get Fig. number with \ref{}
   { \label{fig:AL} 
   The level of feature points obtained in PROB-SLAM is consistent with that of the original SLAM, and there is no reduction in the number of feature points. Thus, the addition of target detection and the corresponding probability map construction thread has little impact on SLAM performance.}
   \end{figure} 

\subsection{Anchor point performance comparison experiment}
In the ORB-SLAM operation scene, the motion of the object, the unclear object to be detected in the image presented in the camera's field of view, the blurred image caused by the rotation of the camera, and the singular angle of view all bring severe challenges to SLAM positioning, and it is easy to locate errors. If it increases, it will even lead to the loss of positioning. The following figure is a typical situation. 

   \begin{figure} [ht]
   \begin{center}
   \begin{tabular}{c} %% tabular useful for creating an array of images 
   \includegraphics[height=5cm]{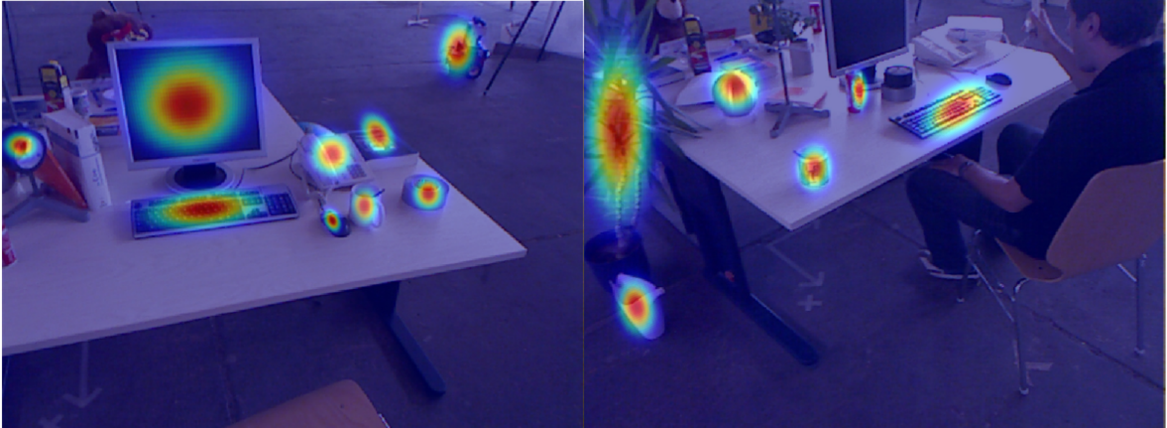}
   \end{tabular}
   \end{center}
   \caption[example] 
%>>>> use \label inside caption to get Fig. number with \ref{}
   { \label{fig:AL} 
   The result is processed by the PROB algorithm. We selected several locations where traditional SALM is prone to significant errors for effect display. Due to the presence of dynamic objects and some unclear fields of view, the original SLAM has significant errors. The experimental results show that the PROB algorithm can effectively alleviate the error problem in these scenarios, and the specific data will be shown later. }   
   \end{figure} 

   \begin{figure} [ht]
   \begin{center}
   \begin{tabular}{c} %% tabular useful for creating an array of images 
   \includegraphics[height=4cm]{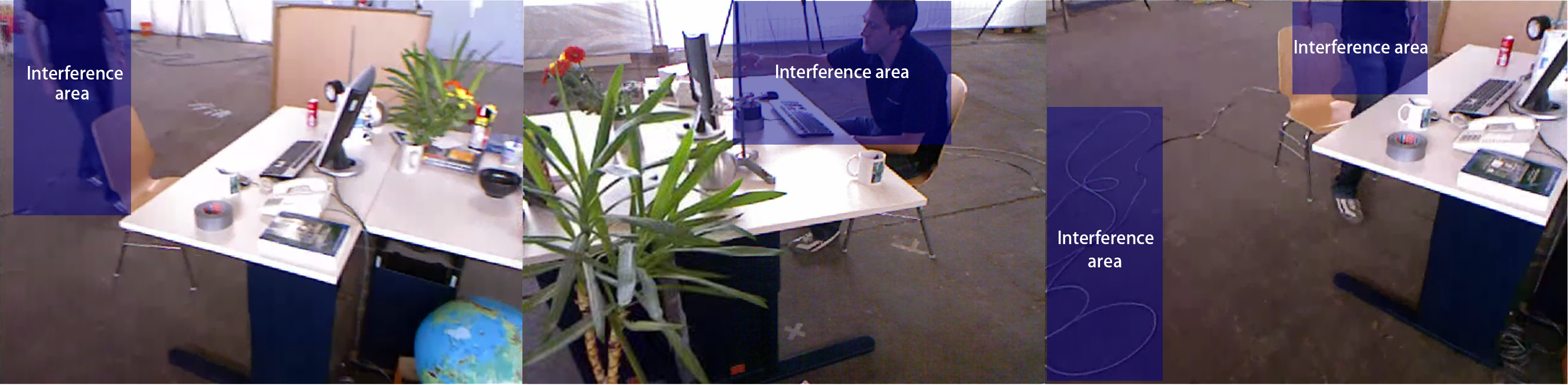}
   \end{tabular}
   \end{center}
   \caption[example] 
%>>>> use \label inside caption to get Fig. number with \ref{}
   { \label{fig:RS} 
   There are dynamic objects moving fast in these scenes, which will cause large errors in some existing slams that use semantic information. At the same time, due to the problem of unclear vision in these scenes, some errors will also be introduced. The PROB-SLAM algorithm in this paper can effectively reduce some errors in these scenarios.}
   \end{figure} 

It is worth noting that there is no performance loss in the scene where many objects are recognized for target detection, and the performance is lower than the original version—many burdens and impacts. However, static SALM cannot discriminate whether the character is moving, resulting in a misjudgment of critical points' dynamic and static attributes. We use the PROB algorithm to optimize critical points based on the target detection probability map, which can well avoid this problem. Through the processing of the PROB algorithm, experiments have proved that the error, in this case, is significantly reduced according to the data response of multiple evaluation indicators. The experimental results fully demonstrate the effectiveness and robustness of the PROB algorithm.

\subsection{Running Error Comparison}
Our experimental results with ORB-SLAM2 show that, unlike the PROB-SLAM algorithm, which has advantages over ORB-SLAM2 in both dynamic and static scenes, the algorithm can achieve a specific improvement in both dynamic and static scenes. The APE and RPE graphs of our algorithm are as follows:

   \begin{figure} [ht]
   \begin{center}
   \begin{tabular}{c} %% tabular useful for creating an array of images 
   \includegraphics[height=6.5cm]{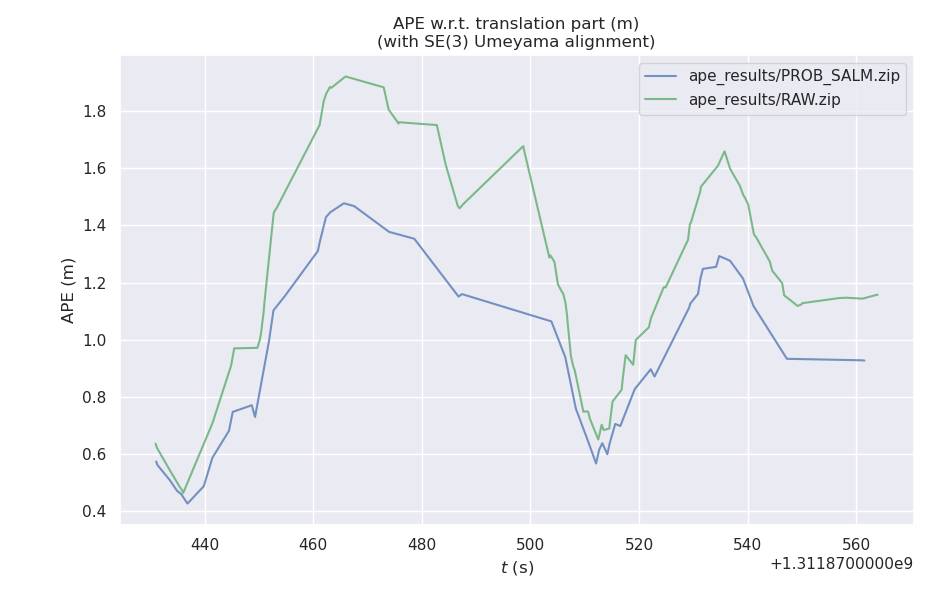}
   \end{tabular}
   \end{center}
   \caption[example] 
%>>>> use \label inside caption to get Fig. number with \ref{}
   { \label{fig:RS} 
   In the line graph, raw represents ORB-SLAM2, and PROB-SLAM represents the system. The two sets of lines correspond to the absolute errors when running the same data set. It can be seen from the figure that when ORB-SLAM2 is at the highest point of error, PROB-SLAM is relatively less than about 15 percent of its error.}
   \end{figure} 

This line chart shows the comparison between the original ORB-SLAM2 and the modified version, where POSE is the version with the probability map added, and POSE-DYNAMIC uses YOLO to remove dynamic objects for predefine knowledge, which shows that in the range of error, It will be improved compared to traditional visual SLAM.

   \begin{figure} [ht]
   \begin{center}
   \begin{tabular}{c} %% tabular useful for creating an array of images 
   \includegraphics[height=6.5cm]{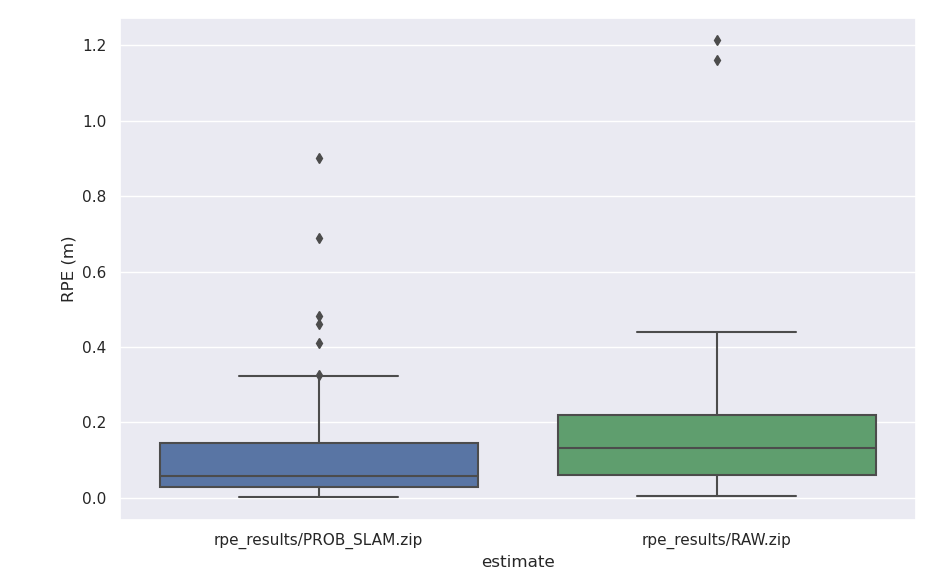}
   \end{tabular}
   \end{center}
   \caption[example] 
%>>>> use \label inside caption to get Fig. number with \ref{}
   { \label{fig:RS} 
   It is the relative error of the two sets of data. It can be seen from the figure that the relative error distribution position of PROB-SLAM is also relatively lower than that of ORB-SLAM2, and the maximum relative error distribution is about 30\% smaller than that of ORB-SLAM2.}
   \end{figure} 

\subsection{Separate module experiments}
To prove the function of our PROB algorithm module, we designed a slice experiment, among them, PROB-SLAM: the algorithm of this paper Only-PROB: direct target detection and convert the target detection image information into a probability map form. 

he results show that the PROB algorithm can directly improve the confidence of the position of crucial points for static objects and can successfully adjust the influence of dynamic objects. It can play a perfect optimization effect in the process of object movement. If Only-PROB adjusts the contrast adjustment index of the probability map too large, it will also cause the loss of positioning.

\subsection{Processing time comparison}
Real-time performance is a significant indicator of the SLAM system, and a more considerable delay will cause a more significant error in SLAM positioning. We tested the average running time of each module, as shown in the table. It includes the time required for ORB-SLAM2 original processing, the time required for PROB-SLAM processing, and the time required for image processing by a separate PROB algorithm. YOLOv5-based semantic threading runs in parallel with ORB feature extraction. The results show that the average processing time per frame of the PROB-SLAM main thread is:

\begin{table}[ht]
   \caption{Processing time of each module.} 
   \label{tab:Paper Margins}
   \begin{center}       
   \begin{tabular}{|l|l|l|l|} 
   \hline
   \rule[-1ex]{0pt}{3ex}  Methods & YOLO & PROB & Tracking \\
   \hline
   \rule[-1ex]{0pt}{3ex}  ORB-SLAM & / & / & 36.5 ms \\
   \hline
   \rule[-1ex]{0pt}{3ex}  PROB-SLAM & 12.28 ms & 0.12 ms & 41.27 ms \\
   \hline
   \end{tabular}
   \end{center}
   \end{table}

\section{Conclusion}
This paper proposes an independent thread for constructing a probability distribution map based on semantic information, which can reduce the impact of semantic information’s uncertainty. Because it is an independent thread, it can be easily integrated into various existing SLAMs. In the system, it will not affect the performance of the original system, so it has a wide range of application scenarios. For example, in some industrial scenarios with fixed particular objects, we can introduce this thread into the existing SLAM system. Accuracy is improved by better utilization of semantic detection information.

% References

\end{document}